\documentclass[runningheads]{llncs}
\usepackage{graphicx}
\usepackage{kotex}
\usepackage[dvipsnames]{xcolor}
\usepackage{multirow}
\usepackage{epsfig}
\usepackage{graphicx}
\usepackage{amsmath}
\usepackage{amssymb}

\usepackage{mathtools}

\usepackage{multirow}

\usepackage[dvipsnames]{xcolor}
\usepackage{booktabs}

\newcommand{\Fref}[1]{Fig.~\ref{#1}}
\newcommand{\Sref}[1]{Sec.~\ref{#1}}
\newcommand{\Tref}[1]{Table~\ref{#1}}

\newcommand{\ie}{\textit{i}.\textit{e}.}
\newcommand{\eg}{\textit{e}.\textit{g}.}
\newcommand{\etal}{\textit{et} \textit{al}.}

\newcommand{\R}{\mathbb{R}}

\usepackage{hyperref}

\begin{document}
\title{Subject Adaptive EEG-based Visual Recognition}
%
%
\author{
Pilhyeon~Lee\inst{1}\and
Sunhee~Hwang\inst{4}\and
Seogkyu~Jeon\inst{1}\and
Hyeran~Byun\inst{1,2,3}\thanks{Corresponding author}}
\authorrunning{P. Lee et al.}
%
\institute{Department of Computer Science, Yonsei University \and
Graduate School of Artificial Intelligence, Yonsei University \and
Graduate Program of Cognitive Science, Yonsei University \and
AI Imaging Tech. Team, LG Uplus\\
\texttt{\{lph1114, jone9312, hrbyun\}@yonsei.ac.kr, sunheehwang@lguplus.co.kr}}
\maketitle              
\begin{abstract}
This paper focuses on EEG-based visual recognition, aiming to predict the visual object class observed by a subject based on his/her EEG signals.
One of the main challenges is the large variation between signals from different subjects.
It limits recognition systems to work only for the subjects involved in model training, which is undesirable for real-world scenarios where new subjects are frequently added.
This limitation can be alleviated by collecting a large amount of data for each new user, yet it is costly and sometimes infeasible.
To make the task more practical, we introduce a novel problem setting, namely \textit{subject adaptive EEG-based visual recognition}.
In this setting, a bunch of pre-recorded data of existing users (source) is available, while only a little training data from a new user (target) are provided.
At inference time, the model is evaluated solely on the signals from the target user.
This setting is challenging, especially because training samples from source subjects may not be helpful when evaluating the model on the data from the target subject.
To tackle the new problem, we design a simple yet effective baseline that minimizes the discrepancy between feature distributions from different subjects, which allows the model to extract subject-independent features.
Consequently, our model can learn the common knowledge shared among subjects, thereby significantly improving the recognition performance for the target subject.
In the experiments, we demonstrate the effectiveness of our method under various settings.
Our code is available at here\footnote[1]{\href{https://github.com/DeepBCI/Deep-BCI/tree/master/1\_Intelligent\_BCI/Subject\_Adaptive\_EEG\_based\_Visual_Recognition}{https://github.com/DeepBCI/Deep-BCI}}.
\end{abstract}

\begin{figure}[t]
  \centering
  \includegraphics[clip=true, width=0.89\textwidth]{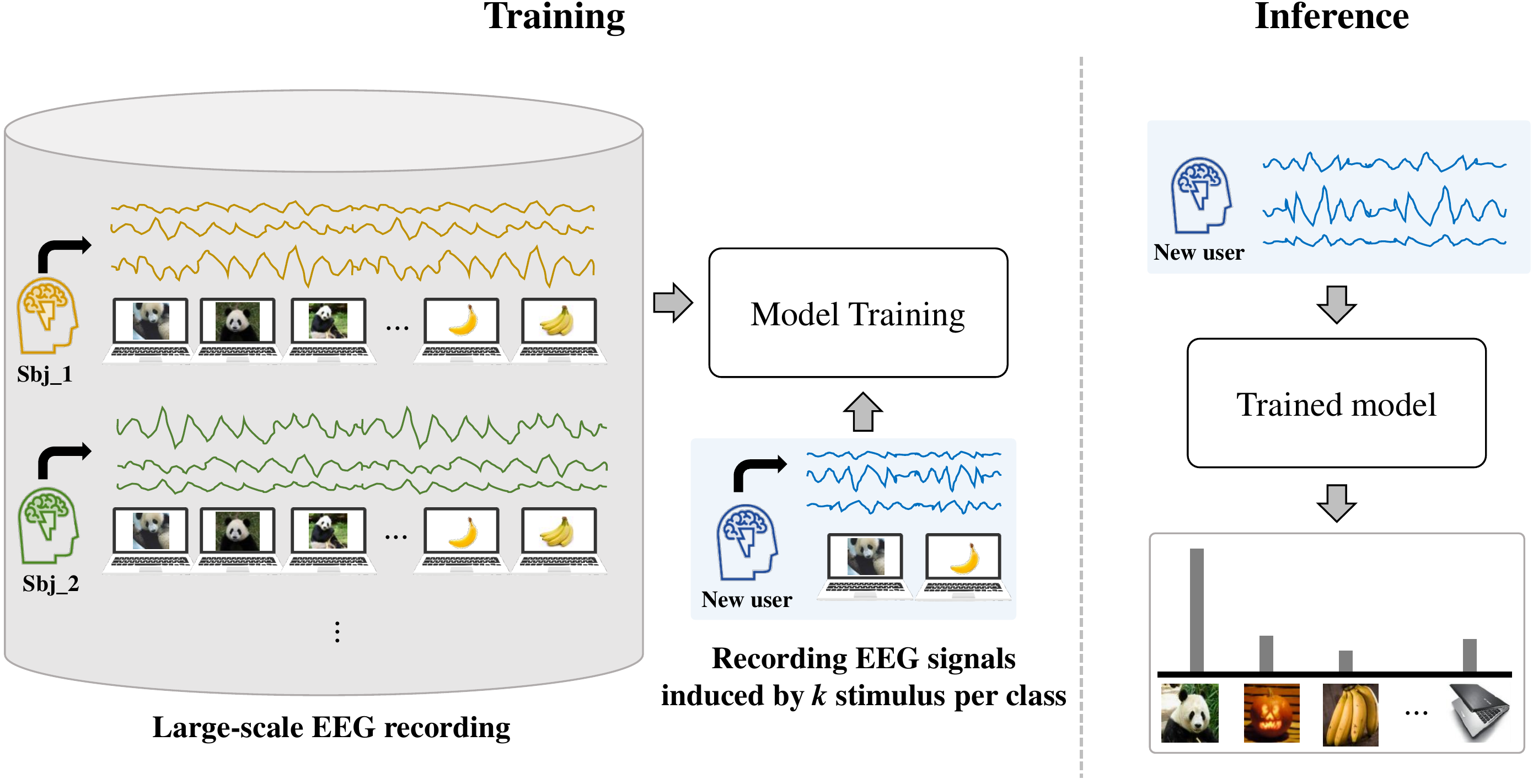}
  \caption{An illustration of \textit{Subject Adaptive EEG-based Visual Recognition}. During the large-scale EEG recording step, abundant sample images are observed by various subjects (source) and we collect their EEG signals. Afterwards, we record EEG signals from a new user~(target) induced by only $k$ stimuli per class. We train the model on the EEG signals from the source and the target subject and expect the trained model to correctly predict the visual classes given unseen EEG signals from the target subject.
  }
  \label{fig:intro_fig}
\end{figure}
\section{Introduction}
Brain-computer interface (BCI) has been a long-standing research topic for decoding human brain activities, playing an important role in reading the human mind with various applications~\cite{bci_app1,bci_app2,bci_app3,bci_app_new}.
For instance, BCI systems enable a user to comfortably control machines without requiring any peripheral muscular activities~\cite{bci_game1,bci_game2}.
In addition, BCI is especially helpful for people suffering from speech or movement disorders, allowing them to freely communicate and express their feelings by thinking~\cite{bci_kb,bci_emotion,bci_wheelchair,vp_eeg_new}.
It also can be utilized to identify abnormal states of brains, such as seizure state, sleep disorder, and dementia~\cite{seizure1,seizure2,dementia,sleep}.
Recently, taking it to the next level, numerous works attempt to decode brain signals for figuring out what audiovisual stimulus is being taken by a person, providing deeper insight for analyzing human perception~\cite{eeg_imgnet,bci_visaud,eeg_music,zsl_sbi}.

There are different ways to collect brain signals, \eg, electroencephalography (EEG), magnetoencephalography (MEG), and functional magnetic resonance imaging (fMRI). 
Among them, EEG is considered the most favorable one to analyze human brain activities since it is non-invasive and promptly acquirable.
With its numerous advantages, EEG-based models have been largely explored by researchers and developed for various research fields such as disorder detection~\cite{disorder1,disorder2}, drowsy detection~\cite{dd_sbi,dd_2}, emotion recognition~\cite{em_sbi,seed_sbd,deap_sbd}, \textit{etc}.

In this paper, we tackle the task of visual recognition based on EEG signals, whose goal is to classify visual stimuli taken by subjects.
Recently, thanks to the effectiveness of deep neural networks (DNNs), existing models have shown impressive recognition performances~\cite{em_sbi,dd_2,eeg_imgnet,eegdnn}.
However, they suffer from the large inter-subject variability of EEG signals, which greatly restricts their scalability.
Suppose that a model faces a new user not included in the training set -- note that this is a common scenario in the real world.
Since the EEG signals from the user are likely to largely differ from those used for training, the model would fail to recognize the classes.
Therefore, in order to retain the performance, it is inevitable to collect EEG signals for training from the new subject, which requires additional costs proportional to the number of the samples.
If we have sufficient training samples for the new subject, the model would show great performance, but it is not the case for the real-world scenario.

To handle this limitation and bypass the expensive cost, we introduce a new practical problem setting, namely \textit{subject adaptive EEG-based visual recognition}.
In this setting, we have access to abundant EEG signals from various source subjects, whereas the signals from a new user (target subject) are scarce, \ie, only a few samples ($k$-shot) are allowed for each visual category.
At inference, the model should correctly classify the EEG signals from the target subject.
\Fref{fig:intro_fig} provides a graphical illustration of the proposed problem setting.

Naturally, involving the copious samples from source subjects in the model training would bring about performance gains compared to the baseline using only signals from the target subject.
However, as aforementioned, the signals obtained from the source and the target subjects are different from each other, and thus the performance improvements are limited.
To maximize the benefits of pre-acquired data from source subjects, we here provide a simple yet effective baseline method.
Our key idea is to allow the model to learn subject-agnostic representations for EEG-based visual recognition.
Technically, together with the conventional classification loss, we design a loss to minimize maximum mean discrepancy (MMD) between feature distributions of EEG signals from different subjects.
On the experiments under a variety of circumstances, our method shows consistent performance improvements over the vanilla method.

Our contributions can be summarized in three-fold.
\begin{itemize}
    \item We introduce a new realistic problem setting, namely subject-adaptive EEG-based visual recognition.
    Its goal is to improve the recognition performance for the target subject whose training samples are limited.
    \item We design a simple baseline method for the proposed problem setting. It encourages the feature distributions between different subjects to be close so that the model learns subject-independent representations.
    \item Through the experiments on the public benchmark, we validate the effectiveness of our model. Specifically, in the extreme 1-shot setting, it achieves the performance gain of 6.4\% upon the vanilla model.
\end{itemize}

\section{Related work}
\label{sec:related_work}
\subsection{Brain activity underlying visual perception}
Over recent decades, research on visual perception has actively investigated to reveal the correlation between brain activity and visual stimuli~\cite{vp_eeg1,vp_eeg2,vp_eeg3}. 
Brain responses induced by visual stimuli come from the occipital cortex that is a brain region for receiving and interpreting visual signals.
In addition, visual information obtained by the occipital lobe is transmitted to nearby parietal and temporal lobes to perceive higher-level information.
Based on this prior knowledge, researchers have tried to analyze brain activities induced by visual stimuli.
Eroğlu~\etal~\cite{vp_eronlu} examine the effect of emotional images with different luminance levels on EEG signals.
They also find that the brightness of visual stimuli can be represented by the activity power of the brain cortex. 
Stewart~\etal~\cite{vp_stewart} attempt to distinguish the presence of visual stimuli within a single trial in EEG recordings.
It is revealed in their analyses that the individual components of EEG signals are spatially located in the visual cortex and are effective in classifying visual states.
More recently, Spampinato~\etal~\cite{eeg_imgnet} tackle the problem of EEG-based visual recognition by learning a discriminative manifold of brain activities on diverse visual categories.
Besides, they build a large-scale EEG dataset for training deep networks and demonstrate that human visual perception abilities can be transferred to deep networks.
Kavasidis~\etal~\cite{brain2image} propose to reconstruct the observed images by decoding EEG signals.
They find that EEG contains some patterns related to visual contents, which can be used to effectively generate images that are semantically coherent to the visual stimuli.

In line with these works, we build a visual recognition model to decode EEG signals induced by visual stimuli.
In addition, we design and tackle a new practical problem setting where a limited amount of data is allowed for new users.

\subsection{Subject-independent EEG-based classification}
Subject-dependent EEG-based classification models have widely been studied, achieving the noticeable performances~\cite{mi1_sbd,mi2_sbd,seed_sbd,deap_sbd,hwang2021bci}. 
However, EEG signal patterns greatly vary among individuals, building a subject-independent model remains an important research topic to be solved.
Hwang~\etal~\cite{em_sbi} train a subject-independent EEG-based emotion recognition model by utilizing an adversarial learning approach to make the model not able to predict the subject labels.
Zhang~\etal~\cite{attention_sbi} propose a convolutional recurrent attention model to classify movement intentions by focusing on the most discriminative temporal periods from EEG signals.
In~\cite{dd_sbi}, an EEG-based drowsy driving detection model is introduced, which is trained in an adversarial manner with gradient reversal layers in order to encourage feature distribution to be close between subjects.

Besides, to eliminate the expensive calibration process for new users, zero-training BCI techniques are introduced which does not require the re-training.
Lee~\etal~\cite{cnn_ztb} try to find the network parameters that generalize well on common features across subjects.
Meanwhile, Grizou \etal~\cite{ztb} propose a zero-training BCI method that controls virtual and robotic agents in sequential tasks without requiring calibration steps for new users.

Different from the works above, we tackle the problem of EEG-based visual recognition.
Moreover, we propose a new problem setting to reduce the cost of acquiring labeled data for new users, as well as introduce a strong baseline.

\section{Dataset}
\label{sec:dataset}

Before introducing the proposed method, we first present the dataset details for experiments.
We use the publicly available large-scale EEG dataset collected by~\cite{eeg_imgnet} that consists of 128-channel EEG sequences lasting for 440 ms from six different subjects (five male and one female).
The EEG signals are filtered using a notch filter (49-51 Hz) and a band-pass filter (14-72 Hz) to include two frequency bands, \ie, Beta and Gamma.
The dataset contains 40 easily distinguishable object categories from ImageNet~\cite{ImageNet}, which are listed in \Tref{table:class_list}.
The number of image samples looked at by subjects is 50 for each class, constituting a total of 2,000 samples.
We use the official splits, keeping the ratio of training, validation, and test sets as 4:1:1.
The dataset contains a total of 6 splits and we measure the mean and the standard deviation of performance of 6 runs in the experiments.
We refer readers to the original paper~\cite{eeg_imgnet} for further details about the dataset.

\begingroup
\setlength{\tabcolsep}{5pt} 
\renewcommand{\arraystretch}{1.2} 

\begin{table*}[t]
\caption{
The list of object classes utilized for collecting EEG signals with ImageNet~\cite{ImageNet} class indices.
}
\centering
\resizebox{1.0\textwidth}{!}{

\begin{tabular}{cc|cc|cc|cc}
\toprule
n02106662 & German shepherd & n02951358 & Canoe & n03445777 & Golf ball & n03888257 & Parachute \\
n02124075 & Egyptian cat & n02992529 & Cellular telephone & n03452741 & Grand piano & n03982430 & Pool table \\
n02281787 & Lycaenid & n03063599 & Coffee mug & n03584829 & Iron & n04044716 & Radio telescope \\
n02389026 & Sorrel & n03100240 & Convertible & n03590841 & Jack-o'-lantern & n04069434 & Reflex camera \\
n02492035 & Capuchin & n03180011 & Desktop computer & n03709823 & Mailbag & n04086273 & Revolver \\
n02504458 & African elephant & n03197337 & Digital watch & n03773504 & Missile & n04120489 & Running shoe \\
n02510455 & Giant panda & n03272010 & Electric guitar& n03775071 & Mitten & n07753592 & Banana \\
n02607072 & Anemone fish & n03272562 & Electric locomotive & n03792782 & Mountain bike & n07873807 & Pizza \\
n02690373 & Airliner & n03297495 & Espresso maker & n03792972 & Mountain tent & n11939491 & Daisy \\
n02906734 & Broom & n03376595 & Folding chair & n03877472 & Pajama & n13054560 & Bolete \\
\bottomrule
\end{tabular}

}
\label{table:class_list}
\end{table*}

\endgroup

\section{Method}
\label{sec:method}
\begin{figure}[t]
  \centering
  \includegraphics[clip=true, width=0.86\textwidth]{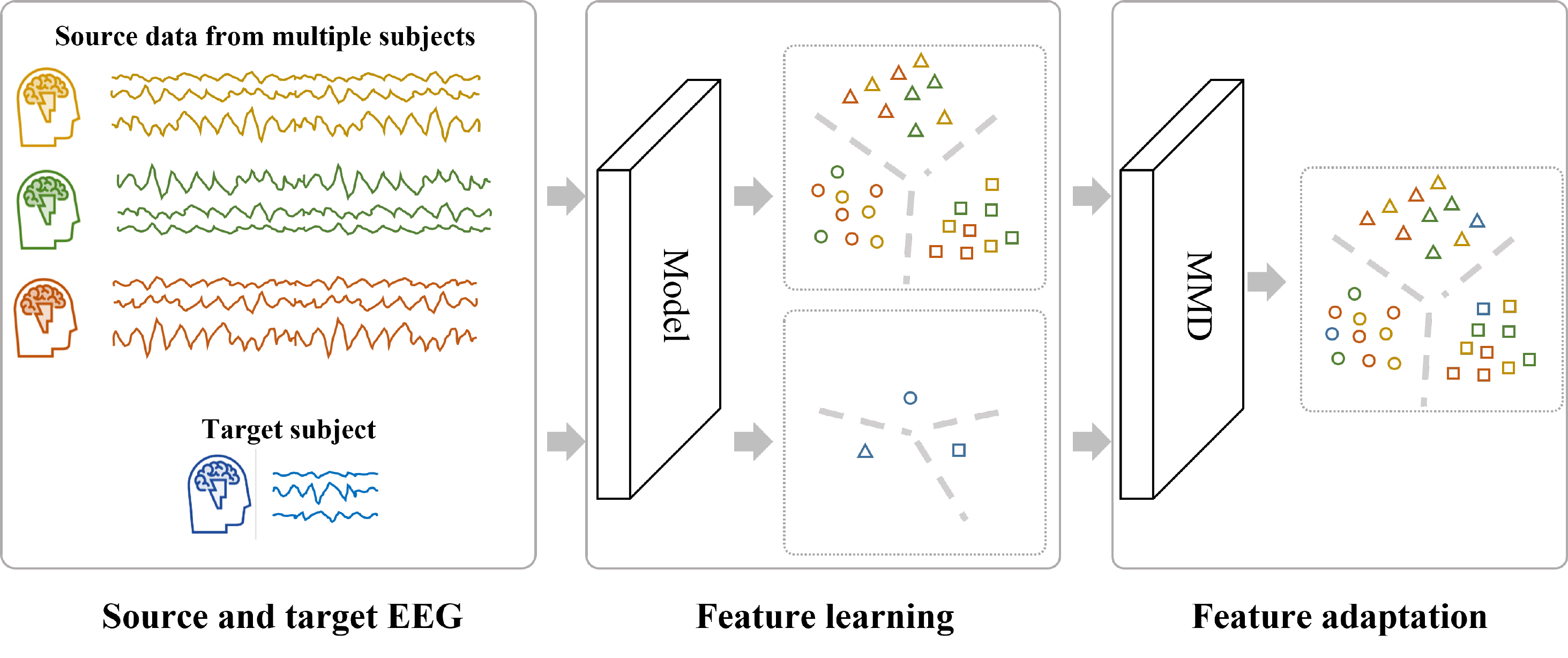}
  \caption{An overview of the proposed method. 
  Colors and shapes respectively represent subject identities and classes. During feature learning, we train the model to accurately predict the class from the EEG signals. To alleviate the feature discrepancy of source and target signals, we propose a feature adaptation stage which minimizes the maximum mean discrepancy. Consequently, both source and target features are projected on the same manifold, enabling accurate predictions on target signals during inference.
  }
  \label{fig:main_architecture}
\end{figure}

In this section, we first define the proposed problem setting~(\Sref{subsec:problem}).
Then, we introduce a baseline method with subject-independent learning to tackle the problem.
Its network architecture is illustrated in \Sref{subsec:architecture}, followed by the detailed subject-independent learning scheme (\Sref{subsec:subject_independent}).
An overview of our method is depicted in \Fref{fig:main_architecture}.

\subsection{Subject Adaptive EEG-based Visual Recognition}
\label{subsec:problem}
We start by providing the formulation of the conventional EEG-based visual recognition task.
Let $\mathcal{D}^{s}=\{(x_i^s, y_i^s)\}_{i=1}^{N^s}$ denote the dataset collected from the $s$-th subject.
Here, $x_i^s \in \R^{D \times T}$ denotes the $i$-th EEG sample of subject $s$ with its channel dimension $D$ and the duration $T$, while $y_i^s \in \R^{K}$ is the corresponding ground-truth visual category observed by the subject and $N^s$ is the number of the samples for subject $s$.
In general, the EEG samples are abundant for each subject, \ie, $N^s \gg 0$.
To train a deep model, multiple datasets from different subjects are assembled to build a single training set $\mathcal{D}=\{ \mathcal{D}^1, \mathcal{D}^2, ... , \mathcal{D}^S \}$, where $S$ is the total number of subjects.
At inference, given an EEG sample $x^{s^\prime}$, the model should predict its category.
Here, it is assumed that the input signal at test time is obtained by one of the subjects whose samples are used during the training stage, \ie, $s^\prime \in [1,S]$.
However, this conventional setting is impractical especially for the case where EEG data from new subjects are scarce.

Instead, we propose a more realistic problem setting, named \textit{Subject Adaptive EEG-based Visual Recognition}.
In this setting, we aim to utilize the knowledge learned from abundant data of source subjects to classify signals from a target subject whose samples are rarely accessible.
For that purpose, we first divide the training set into source and target sets, \ie, $\mathcal{D}_{src}$ and $\mathcal{D}_{trg}$.
We choose a subject and set it to be the target while the rest become the sources.
For example, letting subject $S$ be the target, $\mathcal{D}_{src}=\{ \mathcal{D}^1, \mathcal{D}^2, ... , \mathcal{D}^{S-1} \}$ and $\mathcal{D}_{trg}= \hat{\mathcal{D}}^{S} \subset \mathcal{D}^{S}$.
Based on the sparsity constraint, the target dataset contains only a few examples, \ie, $\hat{\mathcal{D}}^{S} = \{(x_j^S, y_j^S)\}_{j=1}^{\hat{N}^S}$, where $\hat{N}^S \ll N^S$.
In practice, we make the target set have only $k$ samples with their labels per class ($k$-shot).
Note that we here use the $S$-th subject as the target, but any subject can be the target without loss of generality.
After trained on $\mathcal{D}_{src}$ and $\mathcal{D}_{trg}$, the model is supposed to predict the class of an unseen input signal $x^S$ which is obtained from the target subject $S$.

\subsection{Network Architecture}
\label{subsec:architecture}
In this section, we describe the architectural details of the proposed simple baseline method.
Our network is composed of a sequence encoder $f$, an embedding layer $g$, and a classifier $h$.
The sequence encoder $f(\cdot)$ is a single-layer gated recurrent unit~(GRU), which takes as input an EEG sample and outputs the extracted feature representation $z = f(x) \in \R^{D_{seq}}$, where $\R^{D_{seq}}$ is the feature dimension.
Although the encoder produces the hidden representation for every timestamp, we only use the last feature and discard the others since it encodes the information from all timestamps.
Afterwards, the feature $z$ is embedded to the semantic manifold by the embedding layer $g(\cdot)$, \ie, $w = g(z) \in \R^{D_{emb}}$, where $\R^{D_{emb}}$ is the dimension of embedded features.
The embedding layer $g(\cdot)$ is composed of a fully-connected (FC) layer with an activation function.
As the final step, we feed the embedded feature $w$ to the classifier $h(\cdot)$ consisting of a FC layer with the softmax activation, producing the class probability $p(\mathbf{y}|x;\theta) = h(w) \in \R^{K}$. Here, $\theta$ is a set of the trainable parameters in the overall network. To train our network for the classification task, we minimize the cross-entropy loss as follows.
\begin{equation}
    \mathcal{L}_{\text{cls}} = \frac{-1}{|\mathcal{D}_{src}| + |\mathcal{D}_{trg}|}\sum_{(x_i,y_i) \in \mathcal{D}_{src} \cup \mathcal{D}_{trg}}y_i \log p(y_i|x_i;\theta),
    \label{equ:cross_entropy}
\end{equation}
where $|\mathcal{D}_{src}|$ and $|\mathcal{D}_{trg}|$ indicate the number of samples in source and target sets.

\subsection{Subject-independent Feature Learning}
\label{subsec:subject_independent}
In spite of the learned class-discriminative knowledge, the model might not fully benefit from the data of source subjects due to the feature discrepancy from different subjects.
To alleviate this issue and better exploit the source set, we propose a simple yet effective framework, where subject-independent features are learned by minimizing the divergence between feature distributions of source and target subjects.
Concretely, for the divergence metric, we estimate the multi-kernel maximum mean discrepancy~(MK-MMD)~\cite{long2015mmd} between the feature distributions $Z^{s_i}$ and $Z^{s_j}$ from two subjects $s_i$ and $s_j$ as follows.
\begin{equation}
    \text{MMD}(Z^{s_i}, Z^{s_j}) = \left \|\frac{1}{N^{s_i}}\sum^{N^{s_i}}_{n=1}\phi(z^{s_i}_n) - \frac{1}{N^{s_j}}\sum^{N^{s_j}}_{m=1}\phi(z^{s_j}_m)\right \|_{F},
    \label{equ:mmd}
\end{equation}
where $\phi(\cdot):\mathcal{W}\rightarrow\mathcal{F}$ is the mapping function to the reproducing kernel Hilbert space, while $\left\|\cdot\right\|_{F}$ indicates the Frobenius norm.
$z_n^{s_i}$ denotes the $n$-th feature from subject $s_i$ encoded by the sequence encoder $f$, whereas $N^{s_i}$ and $N^{s_j}$ are the total numbers of samples from the $s_i$-th and the $s_j$-th subjects in the training set, respectively.
In practice, we use the samples in an input batch rather than the whole training set due to the memory constraint.
We note that the embedded feature $w_n^i$ could also be utilized to compute the discrepancy, but we empirically find that it generally performs inferior to the case of using $z_n^i$ (\Sref{subsec:location_MMD}).

Reducing the feature discrepancy between different subjects allows the model to learn subject-independent features.
To make feature distributions from all subjects close, we compute and minimize the MK-MMD of all possible pairs of the subjects.
Specifically, we design the discrepancy loss that is formulated as:
\begin{equation}
    \mathcal{L}_{\text{disc}} = \frac{2}{S(S-1)}\sum_{s_i=1}^{S}\sum_{\forall s_j \neq s_i}\text{MMD}(Z^{s_i}, Z^{s_j}),
    \label{equ:mmd_loss}
\end{equation}
where $S$ is the number of the subjects in the training data including the target.

By minimizing the discrepancy loss, our model could learn subject-independent features and better utilize the source data to improve the recognition performance for the target subject.
The overall training loss of our model is a weighted sum of the losses, which is computed as follows:
\begin{equation}
    \mathcal{L}_{\text{total}} = \mathcal{L}_{\text{cls}} + \lambda \mathcal{L}_{\text{disc}},
    \label{equ:overall_loss}
 \end{equation}
where $\lambda$ is the weighting factor, which is empirically set to 1.

\section{Experiments}
\label{sec:experiments}

\subsection{Implementation Details}
\label{subsec:implementation}
The input signals for our method contain a total of 128 channels ($D=128$) with a recording unit of 1~\textit{ms}, each of which lasts for 440~\textit{ms}.
Following \cite{eeg_imgnet}, we only use the signals within the interval of 320-480~\textit{ms}, resulting in the temporal dimension $T=160$.
As described in \Sref{subsec:architecture}, our model consists of a single-layer gated recurrent unit~(GRU) followed by two fully-connected layers respectively for embedding and classification.
For all layers but the classifier, we set their hidden dimensions to the same one with input signals to preserve the dimensionality, \ie, $D_{seq}=D_{emb}=128$.
For non-linearity, we put the Leaky ReLU activation after the embedding layer $g$ with $\alpha=0.2$.
To estimate multi-kernel maximum mean discrepancy, we use the radial basis function (RBF) kernel~\cite{vert2004primer} as the mapping function.
For effective learning, we make sure that all the subjects are included in a single batch.
Technically, we randomly pick 200 examples from each source dataset and take all samples in the target dataset to configure a batch.
Our model is trained in an end-to-end fashion from scratch without pre-training.
For model training, we use the Adam~\cite{kingma2014adam} optimizer with a learning rate of $10^{-3}$.

\begingroup
\setlength{\tabcolsep}{10pt} 
\renewcommand{\arraystretch}{1.1} 

\begin{table*}[t]
\caption{
Quantitative comparison of methods by changing the target subject. For evaluation, we select one subject as a target and set the rest as sources, then compute the top-$k$ accuracy for the test set from the target subject. Note that only a single target sample for each class is included in training, \ie, $1$-shot setting.
We measure the mean and the standard deviation of a total of 5 runs following the official splits.
}
\centering
\resizebox{.84\textwidth}{!}{
\begin{tabular}{c|ccc|ccc}
\toprule
\multicolumn{7}{c}{Validation set} \\ \midrule
\multirow{2}{*}{Subject}  & \multicolumn{3}{c|}{top-1 accuracy (\%)} & \multicolumn{3}{c}{top-3 accuracy (\%)} \\ 
       & $k$-shot      & Vanilla     & Ours   & $k$-shot      & Vanilla     & Ours  \\ \midrule
\#0 & $13.5_{\pm2.1}$   & $29.3_{\pm1.9}$   & $\textbf{35.7}_{\pm1.9}$    & $22.6_{\pm2.8}$   & $51.6_{\pm3.0}$   & $\textbf{58.1}_{\pm2.9}$   \\
\#1 & $12.6_{\pm2.1}$   & $21.8_{\pm2.3}$   & $\textbf{29.0}_{\pm3.6}$    & $22.3_{\pm2.5}$   & $41.0_{\pm5.1}$   & $\textbf{49.5}_{\pm3.5}$   \\
\#2 & $17.0_{\pm1.6}$   & $25.3_{\pm0.9}$   & $\textbf{30.8}_{\pm2.2}$    & $29.8_{\pm2.2}$   & $44.4_{\pm2.1}$   & $\textbf{53.1}_{\pm2.6}$   \\
\#3 & $27.8_{\pm1.7}$   & $28.8_{\pm2.2}$   & $\textbf{31.9}_{\pm3.9}$    & $41.6_{\pm2.1}$   & $47.8_{\pm4.1}$   & $\textbf{52.6}_{\pm3.7}$   \\
\#4 & $16.3_{\pm2.8}$   & $25.9_{\pm1.9}$   & $\textbf{36.2}_{\pm3.3}$    & $25.9_{\pm2.3}$   & $44.4_{\pm2.7}$   & $\textbf{61.0}_{\pm4.7}$   \\
\#5 & $9.2_{\pm1.4}$   & $20.7_{\pm2.9}$   & $\textbf{25.8}_{\pm1.7}$    & $16.9_{\pm2.5}$   & $40.1_{\pm3.9}$   & $\textbf{47.5}_{\pm3.4}$  \\ \midrule

\multicolumn{7}{c}{Test set} \\ \midrule
\multirow{2}{*}{Subject}  & \multicolumn{3}{c|}{top-1 accuracy (\%)} & \multicolumn{3}{c}{top-3 accuracy (\%)} \\ 
       & $k$-shot      & Vanilla     & Ours   & $k$-shot      & Vanilla     & Ours  \\ \midrule
\#0 & $12.2_{\pm2.1}$   & $24.3_{\pm0.9}$   & $\textbf{29.6}_{\pm4.9}$    & $20.4_{\pm2.5}$   & $48.3_{\pm2.3}$   & $\textbf{56.8}_{\pm4.1}$   \\
\#1 & $10.3_{\pm2.2}$   & $18.1_{\pm2.7}$   & $\textbf{25.4}_{\pm2.4}$    & $20.8_{\pm2.1}$   & $39.0_{\pm1.9}$   & $\textbf{49.0}_{\pm2.4}$   \\
\#2 & $15.5_{\pm2.9}$   & $23.9_{\pm3.0}$   & $\textbf{29.2}_{\pm3.7}$    & $29.9_{\pm3.4}$   & $44.3_{\pm4.3}$   & $\textbf{54.5}_{\pm3.1}$   \\
\#3 & $26.2_{\pm3.2}$   & $27.4_{\pm3.2}$   & $\textbf{32.1}_{\pm4.3}$    & $41.7_{\pm3.9}$   & $47.9_{\pm4.2}$   & $\textbf{53.6}_{\pm4.0}$   \\
\#4 & $15.2_{\pm1.9}$   & $22.7_{\pm1.2}$   & $\textbf{35.3}_{\pm3.6}$    & $24.5_{\pm2.0}$   & $44.8_{\pm3.5}$   & $\textbf{60.7}_{\pm4.9}$   \\
\#5 & $7.0_{\pm1.0}$   & $18.9_{\pm2.9}$   & $\textbf{21.4}_{\pm2.6}$    & $15.3_{\pm1.8}$   & $38.4_{\pm4.1}$   & $\textbf{45.0}_{\pm4.1}$  \\
    \bottomrule
\end{tabular}

}
\label{table:quant_subject}
\end{table*}

\endgroup

\subsection{Quantitative Results}
\label{subsec:quantitative}
To validate the effectiveness of our method, we compare it with two different competitors: $k$-shot baseline and the vanilla model.
First, the $k$-shot method is trained exclusively on the target dataset.
As the amount of target data is limited, the model is expected to poorly perform and it would serve as the baseline for investigating the benefit of source datasets.
Next, the vanilla model is a variant of our method that discards the discrepancy loss.
Its training depends solely on the classification loss without considering subjects, and thus it can demonstrate the effect of abundant data from other unrelated subjects.

\begingroup
\setlength{\tabcolsep}{10pt} 
\renewcommand{\arraystretch}{1.1} 

\begin{table*}[t]
\caption{
Quantitative comparison of methods by changing the number of target samples per class provided during training.
The value of $k$ means that only $k$ samples of the target subject are used for training.
We measure the mean and the standard deviation of a total of 5 runs for all subjects following the official splits.
}
\centering
\resizebox{.82\textwidth}{!}{
\begin{tabular}{c|ccc|ccc}
\toprule
\multicolumn{7}{c}{Validation set} \\ \midrule
\multirow{2}{*}{$k$}  & \multicolumn{3}{c|}{top-1 accuracy (\%)} & \multicolumn{3}{c}{top-3 accuracy (\%)} \\ 
       & $k$-shot      & Vanilla     & Ours   & $k$-shot      & Vanilla     & Ours  \\ \midrule
1 & $16.0_{\pm0.6}$   & $25.3_{\pm1.0}$   & $\textbf{31.7}_{\pm1.5}$    & $26.5_{\pm0.9}$   & $44.9_{\pm1.3}$   & $\textbf{53.6}_{\pm1.9}$   \\
2 & $33.2_{\pm1.2}$   & $41.7_{\pm1.9}$   & $\textbf{46.3}_{\pm1.8}$    & $50.1_{\pm1.0}$   & $65.2_{\pm2.0}$   & $\textbf{70.2}_{\pm1.6}$   \\
3 & $49.9_{\pm0.4}$   & $54.4_{\pm1.0}$   & $\textbf{58.9}_{\pm0.7}$    & $68.5_{\pm0.7}$   & $77.6_{\pm0.7}$   & $\textbf{80.8}_{\pm1.2}$   \\
4 & $61.9_{\pm2.0}$   & $64.6_{\pm1.5}$   & $\textbf{67.5}_{\pm1.2}$    & $79.6_{\pm1.7}$   & $85.1_{\pm1.1}$   & $\textbf{86.8}_{\pm1.2}$   \\
5 & $70.0_{\pm1.6}$   & $72.0_{\pm1.3}$   & $\textbf{73.5}_{\pm1.1}$    & $85.6_{\pm1.7}$   & $89.6_{\pm0.9}$   & $\textbf{90.0}_{\pm1.0}$  \\ \midrule
\multicolumn{7}{c}{Test set} \\ \midrule
\multirow{2}{*}{$k$}  & \multicolumn{3}{c|}{top-1 accuracy (\%)} & \multicolumn{3}{c}{top-3 accuracy (\%)} \\ 
       & $k$-shot      & Vanilla     & Ours   & $k$-shot      & Vanilla     & Ours  \\ \midrule
1 & $14.4_{\pm1.6}$   & $22.5_{\pm0.8}$   & $\textbf{28.8}_{\pm1.2}$    & $25.4_{\pm1.8}$   & $43.8_{\pm1.6}$   & $\textbf{53.3}_{\pm1.9}$   \\
2 & $31.2_{\pm1.2}$   & $39.9_{\pm2.0}$   & $\textbf{43.8}_{\pm1.4}$    & $49.3_{\pm2.0}$   & $65.1_{\pm2.1}$   & $\textbf{69.5}_{\pm1.4}$   \\
3 & $48.2_{\pm2.6}$   & $52.6_{\pm1.7}$   & $\textbf{56.4}_{\pm1.7}$    & $67.2_{\pm1.7}$   & $77.0_{\pm1.5}$   & $\textbf{80.4}_{\pm1.1}$   \\
4 & $60.4_{\pm0.9}$   & $62.4_{\pm1.7}$   & $\textbf{64.7}_{\pm1.6}$    & $79.4_{\pm1.1}$   & $84.3_{\pm0.9}$   & $\textbf{85.9}_{\pm1.1}$   \\
5 & $68.1_{\pm1.6}$   & $69.5_{\pm1.1}$   & $\textbf{70.1}_{\pm1.0}$    & $85.6_{\pm1.3}$   & $89.0_{\pm0.5}$   & $\textbf{89.2}_{\pm0.5}$  \\
    \bottomrule
\end{tabular}

}
\label{table:quant_k}
\end{table*}

\endgroup

\paragraph{Comparison in the 1-shot setting.}
We first explore the most extreme scenario of our subject adaptive EEG-based visual classification, \ie, the 1-shot setting.
In this setting, only a single example for each visual category is provided for the target subject.
The experimental results are summarized in \Tref{table:quant_subject}.
As expected, the $k$-shot baseline performs the worst due to the scarcity of training data.
When including the data from source subjects, the vanilla setting improves the performance to an extent.
However, we observe that the performance gain is limited due to the representation gap between subjects.
On the other hand, our model manages to learn subject-independent information and brings a large performance boost upon the vanilla method without regard to the choice of the target subject.
Specifically, the top-1 accuracy of subject \#1 on the validation set is improved by 7.2\% from the vanilla method.
This clearly validates the effectiveness of our approach.

\begingroup
\setlength{\tabcolsep}{10pt} 
\renewcommand{\arraystretch}{1.1} 

\begin{table*}[t]
\caption{
Ablation on the location of feature adaptation.
We compare two variants that minimize discrepancy after the sequence encoder $f$ and the embedding layer $g$, respectively.
We measure the mean and the standard deviation of a total of 5 runs for all subjects.
}
\centering
\resizebox{.56\textwidth}{!}{
\begin{tabular}{c|cc|cc}
\toprule
\multirow{2}{*}{$k$}  & \multicolumn{2}{c|}{top-1 accuracy (\%)} & \multicolumn{2}{c}{top-3 accuracy (\%)} \\ 
       & after $f$     & after $g$   & after $f$      & after $g$     \\ \midrule
1 & $31.7_{\pm1.5}$   & $\textbf{32.4}_{\pm0.7}$    & $53.6_{\pm1.9}$   & $\textbf{54.8}_{\pm1.1}$   \\
2 & $\textbf{46.3}_{\pm1.8}$   & $46.0_{\pm1.8}$    & $\textbf{70.2}_{\pm1.6}$   & $69.6_{\pm1.9}$   \\
3 & $\textbf{58.9}_{\pm0.7}$   & $58.3_{\pm1.3}$    & $\textbf{80.8}_{\pm1.2}$   & $80.4_{\pm1.3}$   \\
4 & $\textbf{67.5}_{\pm1.2}$   & $65.6_{\pm1.5}$    & $\textbf{86.8}_{\pm1.2}$   & $86.0_{\pm0.9}$   \\
5  & $\textbf{73.5}_{\pm1.1}$   & $72.3_{\pm1.3}$    & $\textbf{90.0}_{\pm1.0}$   & $89.7_{\pm0.7}$  \\ \bottomrule
\end{tabular}

}
\label{table:mmd_location}
\end{table*}

\endgroup

\paragraph{Comparison with varying $k$.}
To investigate the performance in diverse scenarios, we evaluate the models with varying $k$ for the $k$-shot setting.
Specifically, we change $k$ from 1 to 5 and the results are provided in \Tref{table:quant_k}.
Obviously, increasing $k$ leads to performance improvements for all the methods.
On the other hand, it can be also noticed that regardless of the choice of $k$, our method consistently outperforms the competitors with non-trivial margins, indicating the efficacy and the generality of our method.
Meanwhile, the performance gaps between the methods get smaller as $k$ grows, since the benefit of source datasets vanishes as the volume of the target dataset increases.
We note, however, that a large value of $k$ is impractical and sometimes even unreachable in the real-world setting.

\subsection{Analysis on the location of feature adaptation}
\label{subsec:location_MMD}
Our feature adaptation with the discrepancy loss (Eq. \ref{equ:mmd_loss}) can be adopted into any layer of the model.
To analyze the effect of its location, we compare two variants that minimize the distance of feature distributions after the sequence encoder $f$ and the embedding layer $g$, respectively.
The results are shown in \Tref{table:mmd_location}, where the variant ``after~$f$'' generally shows better performance compared to ``after~$g$'' except for the case where $k$ is set to 1.
We conjecture that this is because it is incapable for a single GRU encoder (\ie, $f$) to align feature distributions from different subjects well when the amount of the target dataset is too small.
However, with a sufficiently large $k$, the variant ``after~$f$'' consistently performs better with obvious margins.
Based on these results, we compute the MK-MMD on the features after the sequential encoder $f$ by default.

\section{Concluding Remarks}
\label{sec:conclusion}
In this paper, we introduce a new setting for EEG-based visual recognition, namely \textit{subject adaptive EEG-based visual recognition}, where plentiful data from source subjects and sparse samples from a target subject are provided for training.
This setting is cost-effective and practical in that it is often infeasible to acquire sufficient samples for a new user in the real-world scenario.
Moreover, to better exploit the abundant source data, we introduce a strong baseline that minimizes the feature discrepancy between different subjects.
In the experiments with various settings, we clearly verify the effectiveness of our method compared to the vanilla model.
We hope this work would trigger further research under realistic scenarios with data scarcity, such as subject generalization~\cite{ghifary2015domain,jeon2021stylizationDG}.

\section*{Acknowledgments}
This work was supported by Institute for Information \& Communications Technology Planning \& Evaluation (IITP) grant funded by the Korea government (MSIT) (No. 2017-0-00451: Development of BCI based Brain and Cognitive Computing Technology for Recognizing Users Intentions using Deep Learning, No. 2020-0-01361: Artificial Intelligence Graduate School Program (YONSEI UNIVERSITY)).

\bibliographystyle{splncs04}
\bibliography{egbib}
\end{document}